\documentclass[a4paper]{article}

\usepackage{INTERSPEECH2020}
\usepackage{listings}
\usepackage{hyperref, multicol, multirow, subcaption}
\graphicspath{ {img/} } 

\title{BlaBla: Linguistic Feature Extraction for Clinical Analysis in Multiple Languages
}
\name{Abhishek Shivkumar, Jack Weston, Raphael Lenain, Emil Fristed}
\address{Novoic Ltd}
\email{\{abhishek, jack, raphael, emil\}@novoic.com}

\begin{document}

\maketitle
\begin{abstract}

We introduce BlaBla, an open-source Python library for extracting linguistic features with proven clinical relevance to neurological and psychiatric diseases across many languages. BlaBla is a unifying framework for accelerating and simplifying clinical linguistic research. The library is built on state-of-the-art NLP frameworks and supports multithreaded/GPU-enabled feature extraction via both native Python calls and a command line interface. We describe BlaBla’s architecture and clinical validation of its features across 12 diseases. We further demonstrate the application of BlaBla to a task visualizing and classifying language disorders in three languages on real clinical data from the AphasiaBank dataset. We make the codebase freely available to researchers with the hope of providing a consistent, well-validated foundation for the next generation of clinical linguistic research.

\end{abstract}
\noindent\textbf{Index Terms}: Alzheimer's disease, aphasia, BlaBla, computational linguistics, healthcare, machine learning (ML), natural language processing (NLP), Novoic, speech and language disorders, speech processing.

\section{Introduction} 

Natural language processing (NLP) has many diverse use cases, including machine translation \cite{sutskever2014sequence, bahdanau2014neural}, question answering \cite{devlin2018bert}, named entity recognition \cite{nadeau2007survey}, sentiment analysis \cite{pang2008opinion}, document summarization \cite{zhang2019hibert} and topic modelling \cite{blei2012probabilistic}. Some of the most important applications lie in the medical domain, where NLP is applied to tasks such as automatic processing of electronic health records (EHRs) \cite{ohno2011realizing}. Feature extraction from language (e.g. the number of noun phrases) is also used directly for analyzing patterns of speech to characterize different medical conditions. Speech and language changes have been observed in neurodegenerative diseases (e.g. Alzheimer's disease), motor conditions (e.g. Parkinson's) and affective disorders (e.g. depression) \cite{boschi2017connected,low2020automated,chan2019speech}; see Table \ref{tab:table1} for details. Spoken language might therefore offer a universal and accessible means for measuring neurological health, but its application has been so far limited: feature extraction from language is often done manually or using a patchwork of different packages and custom scripts \cite{de2013predicting,fraser2016linguistic}. Further, most work to date has been done in the English language. The research community is therefore faced with a pressing need to harmonize the extraction of language features and generalize this to languages other than English.

We introduce BlaBla, an easy-to-use Python library for language feature extraction, written with the aim of providing a unifying and consistent implementation of commonly used language features from the clinical literature across multiple languages. We write BlaBla with the aim of equipping every member of the NLP community (including those without prior Python knowledge) with powerful tooling by designing two distinct ways to extract features: either by calling BlaBla programmatically in Python or via its command line interface (CLI). BlaBla was built to facilitate fast prototyping and complement current machine learning workflows using sklearn, TensorFlow or PyTorch for example.

Modern linguistic analysis, especially that in the clinical domain, has previously required stitching together multiple tools and resources. This typically includes transcript-parsing software such as CLAN \cite{macwhinney2010transcribing} and NXT \cite{calhoun2010nxt}, low-level NLP libraries such as NLTK \cite{loper2002nltk}, Stanford CoreNLP \cite{manning2014stanford}, and hand-engineered features like Honor\'e's statistic \cite{honore1979some} and mean Yngve depth \cite{yngve1972depth} following a literature review. Performing such analyses in languages other than, or in addition to, English further complicates the task. What lacks is a single source of truth able to operate faithfully across multiple languages.

Today, a growing number of open-source NLP toolkits with sophisticated offerings exist, with examples including CoreNLP \cite{manning2014stanford}, FLAIR \cite{akbik2019flair} and spaCy \cite{spacy}. A major limitation of these toolkits has been support of only a few languages: CoreNLP (implemented  in Java) supports 6 human languages, FLAIR (Python) 12, and spaCy (Python) 10 \cite{qi2020stanza}. Universal Dependencies (UD) is a framework for consistent annotation of grammar (parts of speech, morphological features and syntactic dependencies) across a variety of human languages \cite{nivre2016universal}. The UD community has built a large database of treebanks in many languages, now used for training some newer toolkits. One example is UDPipe \cite{straka2018udpipe}, a C++ toolkit with support for 61 human languages. Very recently, Stanza was released by the Stanford NLP group \cite{qi2020stanza}, a Python library with a fully neural pipeline trained on UD treebanks and other multilingual corpora. Based on raw text input, it produces annotations including tokenization, multi-word token expansion, lemmatization, part-of-speech (POS) and morphological feature tagging, dependency parsing, and named entity recognition \cite{qi2020stanza}. It further offers a Python interface to CoreNLP, providing additional annotations such as the constituency parse tree, though only in the 6 languages supported by CoreNLP. 

BlaBla is built using Stanza, which we consider the state-of-the-art NLP toolkit, and CoreNLP. Our algorithms sit on top of these packages to form a unified library for extracting clinically relevant features:
\begin{itemize}
    \item \textbf{Phonetic and phonological features} calculated using timestamp-aligned transcripts.
    \item \textbf{Lexicosemantic features} derived from Stanza's dependency parse tree.
    \item \textbf{Morphosyntactic and syntactic features} derived from CoreNLP's constituency parse tree.
    \item \textbf{Discourse and pragmatic features} derived from Stanza's dependency parse tree.
\end{itemize}
We describe BlaBla's implementation and interface along with the clinical rationale behind its features, then illustrate the ability to rapidly prototype ML models using BlaBla. We take as an example the task of visualizing and classifying aphasic patients in the AphasiaBank dataset \cite{forbes2012aphasiabank} across three languages. We provide a list of reference values on the AMI dataset \cite{carletta2005ami} and the clinical rationale underlying each feature. Finally, we release the BlaBla codebase\footnote{\url{https://github.com/novoic/blabla}} to the research community under an open-source license, along with notebooks containing all the code used in this paper\footnote{\url{https://github.com/novoic/blabla-IS2020}}.




\begin{table*}[!h]
\vspace{0.4cm}
\caption{A description of BlaBla features including their clinical rationale. This includes a set of reference values calculated using BlaBla v0.1 on the AMI corpus. The languages supported by CoreNLP and Stanza are described in the main text (see Section \ref{section:architecture}). The right half of the table is adapted from \cite{boschi2017connected,low2020automated,tomik2010dysarthria,noffs2018speech, chan2019speech}, summarizing clinical validation of recent review papers across indications. $\uparrow$ = the feature increases relative to healthy controls; $\downarrow$ = the feature decreases relative to healthy controls; * = the review has confirmed that the feature does not change; - = unknown. NF/av = nonfluent agrammatic variant primary progressive aphasia (PPA); Sv = semantic variant PPA; L/Pv = logopenic variant PPA; AD = Alzheimer's disease; MND = motor neurone disease, synonymous with amyotrophic lateral sclerosis (ALS); PD = Parkinson's disease; HD = Huntington's disease; MS = multiple sclerosis; MDD = major depressive disorder; HpM = hypomania; Szo = schizophrenia. A dagger ($\dagger$) indicates that the feature expected to scale with the text length so the reference value (if provided) is given only for completeness. Phonetic and phonological features are calculated directly using timestamp-aligned transcripts so their implementation is language-agnostic.} 
\centering 
\resizebox{\textwidth}{!}{
\begin{tabular}{| c| c | c|| c c c c c c c c c c c c |}
\hline\hline
Feature & Base (\#langs) & AMI reference & AD & Sv & L/Pv & NF/av & MND & PD & LDB & HD & MS & MDD & HpM & Szo\\ [0.5ex] 
\hline

 \textbf{Phonetic \& phonological features} &   & &    &  &  &  &  &  &  & & & & & \\
\hline
 Number of pauses$\dagger$ & - & - & - & - & - & - & - & $\uparrow$\cite{boschi2017connected} & -  & - & $\uparrow$\cite{noffs2018speech} & - & - & -\\

Total pause time$\dagger$ & - & - & *\cite{boschi2017connected} & $\uparrow$\cite{boschi2017connected} & - & $\uparrow$\cite{boschi2017connected} & $\uparrow$\cite{boschi2017connected} & *\cite{boschi2017connected} & $\uparrow$\cite{boschi2017connected} & - & - & $\uparrow$\cite{low2020automated} & $\uparrow$\cite{low2020automated} & $\uparrow$\cite{low2020automated}\\

Mean pause duration & - & - & *\cite{boschi2017connected} & $\downarrow$\cite{boschi2017connected} & - & - & - & - & - & - & $\uparrow$\cite{noffs2018speech} & *\cite{low2020automated} & $\uparrow$\cite{low2020automated} & $\uparrow$\cite{low2020automated}\\

Between-utterance pause duration & - & - & - & - & - & - & - & $\uparrow$\cite{boschi2017connected} & - & - & - & - & - & -\\

Hesitation ratio & - & - & $\uparrow$\cite{boschi2017connected} & - & - & $\uparrow$\cite{boschi2017connected} & - & - & - & - & - & - & - & -\\

Speech rate & - & - & $\downarrow$\cite{boschi2017connected} & $\downarrow$\cite{boschi2017connected} & $\downarrow$\cite{boschi2017connected} & $\downarrow$\cite{boschi2017connected} & *\cite{boschi2017connected} & *\cite{boschi2017connected} & $\downarrow$\cite{boschi2017connected} & - & $\downarrow$\cite{noffs2018speech} & $\downarrow$\cite{low2020automated} & - & $\downarrow$\cite{low2020automated}\\

Maximum speech rate & - & - & - & - & $\downarrow$\cite{boschi2017connected} & $\downarrow$\cite{boschi2017connected} & - & - & - & - & - & - & - & -\\

Total phonation time$\dagger$ & - & - & *\cite{boschi2017connected} & - & - & - & - & - & - & - & $\downarrow$\cite{noffs2018speech} & - & - & $\downarrow$\cite{low2020automated}\\

Standardized phonation time$\dagger$ & - & - & $\downarrow$\cite{boschi2017connected} & - & - & - & - & - & - & - & - & - & - & -\\

Total locution time$\dagger$ & - & - & *\cite{boschi2017connected} & *\cite{boschi2017connected} & *\cite{boschi2017connected} & $\uparrow$\cite{boschi2017connected} & *\cite{boschi2017connected} & *\cite{boschi2017connected} & *\cite{boschi2017connected} & - & - & - & - & -\\ \hline

 \textbf{Lexicosemantic features} &   & &  &  &  &  &  &  &  &  &  &  & & \\
\hline
Noun rate & Stanza (66) &  $0.123\pm0.0223$ & *\cite{boschi2017connected} & $\downarrow$\cite{boschi2017connected} & *\cite{boschi2017connected} & *\cite{boschi2017connected} & *\cite{boschi2017connected} & *\cite{boschi2017connected} & *\cite{boschi2017connected} & - & - & - & - & -\\

Verb rate & Stanza (66) & $0.108\pm0.0168$ & $\uparrow$\cite{boschi2017connected} &  *\cite{boschi2017connected} & *\cite{boschi2017connected} & *\cite{boschi2017connected} & *\cite{boschi2017connected} & - & - & *\cite{boschi2017connected} & - & - & - & -\\

Demonstrative rate & Stanza (66) & $0.0263\pm0.00869$ & *\cite{boschi2017connected} & $\uparrow$\cite{boschi2017connected} & - & $\uparrow$\cite{boschi2017connected} & - & - & - & - & - & - & - & -\\

Adjective rate & Stanza (66) & $0.0479\pm0.0116$& *\cite{boschi2017connected} & *\cite{boschi2017connected} & - & *\cite{boschi2017connected} & - & - & - & *\cite{boschi2017connected} & - & - & - & -\\

Pronoun rate & Stanza (66) & $0.144\pm0.0201$ & *\cite{boschi2017connected} & $\uparrow$\cite{boschi2017connected} & $\uparrow$\cite{boschi2017connected} & *\cite{boschi2017connected} & *\cite{boschi2017connected} & - & - & - & - & - & - & -\\

Adverb rate & Stanza (66) & $0.0742\pm0.0157$ & *\cite{boschi2017connected} & $\uparrow$\cite{boschi2017connected} & - & *\cite{boschi2017connected} & - & - & - & *\cite{boschi2017connected} & - & - & - & -\\

Conjunction rate & Stanza (66) & $0.0339\pm0.00946$  & *\cite{boschi2017connected} & - & - & - & - & - & - & - & - & - & - & -\\

Possessive rate & Stanza (66) & $0.192\pm0.0215$ & - & *\cite{boschi2017connected} & - & $\downarrow$\cite{boschi2017connected} & - & - & - & - & - & - & - & -\\

Noun-verb ratio & Stanza (66) & $1.17\pm0.313$ & - & $\downarrow$\cite{boschi2017connected} & - & *\cite{boschi2017connected} & - & - & - & - & - & - & - & -\\

Noun ratio & Stanza (66) & $0.531\pm0.0621$ & *\cite{boschi2017connected} & $\downarrow$\cite{boschi2017connected} & - & *\cite{boschi2017connected} & - & - & - & - & - & - & - & -\\

Pronoun-noun ratio & Stanza (66) & $1.22\pm0.362$ & $\uparrow$\cite{boschi2017connected} & - & - & - & - & - & - & - & - & - & - & -\\

Closed-class word rate & Stanza (66) & $0.312\pm0.0244$ & $\uparrow$\cite{boschi2017connected} & - & - & - & - & - & - & - & - & - & - & -\\

Open-class word rate & Stanza (66) & $0.353\pm0.0315$ & *\cite{boschi2017connected} & $\downarrow$\cite{boschi2017connected} & $\downarrow$\cite{boschi2017connected} & *\cite{boschi2017connected} & - & *\cite{boschi2017connected} & *\cite{boschi2017connected} & - & - & - & - & -\\

Content density & Stanza (66) & $1.14\pm0.122$ & *\cite{boschi2017connected} & $\downarrow$\cite{boschi2017connected} & - & - & - & - & - & - & - & - & - & -\\

Idea density & Stanza (66) & $0.324\pm0.0310$ & $\downarrow$\cite{boschi2017connected} & - & - & - & - & - & - & - & - & - & - & -\\

Honor\'e's statistic & Stanza (66) & $(1.61\pm0.156)\times10^3$ & *\cite{boschi2017connected} & - & - & - & - & - & - & - & - & - & - & -\\

Brunet's index & Stanza (66) & $14.9\pm1.55$ & *\cite{boschi2017connected} & - & - & - & - & - & - & - & - & - & - & -\\

Type-token ratio & Stanza (66) & $0.247\pm0.0886$ & *\cite{boschi2017connected} & *\cite{boschi2017connected} & - & *\cite{boschi2017connected} & - & - & -  & - & - & - & - & -\\

Word length & Stanza (66) & $3.42\pm0.187$& *\cite{boschi2017connected} & $\downarrow$\cite{boschi2017connected} & - & $\downarrow$\cite{boschi2017connected} & - & -  & -  & - & - & - & -  & -\\ \hline

 \textbf{Morphosyntactic \& syntactic features} &   &  &  &  &  &  &  &  &  & & & & & \\
\hline
 Proportion of inflected verbs & Stanza (66) & $0.337\pm0.0930$ & *\cite{boschi2017connected} & *\cite{boschi2017connected} & *\cite{boschi2017connected} & *\cite{boschi2017connected} & *\cite{boschi2017connected} & *\cite{boschi2017connected} & *\cite{boschi2017connected} & - & - & - & - & -\\

Proportion of auxiliary verbs & Stanza (66) & $0.798\pm0.205$ & *\cite{boschi2017connected} & - & - & *\cite{boschi2017connected} & - & - & - & - & - & - & - & -\\

Proportion of gerund verbs & Stanza (66) & $0.0393\pm0.0307$ & $\downarrow$\cite{boschi2017connected} & - & - & - & - & - & - & - & - & - & - & -\\

Proportion of participles & Stanza (66) & $0.141\pm0.0669$ & $\downarrow$\cite{boschi2017connected} & - & - & - & - & - & - & - & - & - & - & -\\

Number of clauses$\dagger$ & CoreNLP (6) & $311\pm250$ & *\cite{boschi2017connected} & *\cite{boschi2017connected} & - & *\cite{boschi2017connected} & - & - & - & *\cite{boschi2017connected} & - & - & - & -\\

Clause rate & CoreNLP (6) & $3.38\pm1.28$& - & *\cite{boschi2017connected} & - & $\downarrow$\cite{boschi2017connected} & - & - & - & - & - & - & - & -\\

Proportion of nouns with determiners & Stanza (66) & $0.544\pm0.0999$ & *\cite{boschi2017connected} & *\cite{boschi2017connected} & *\cite{boschi2017connected} & $\downarrow$\cite{boschi2017connected} & - & - & - & - & - & - & - & -\\

Proportion of nouns with adjectives & Stanza (66) & $0.215\pm0.0635$ & - & - & - & - & - & - & - & - & - & - & - & -\\

Number of noun phrases$\dagger$ & CoreNLP (6) & $450\pm342$ & $\downarrow$\cite{boschi2017connected} & - & - & $\uparrow$\cite{boschi2017connected} & - & - & - & - & - & - & - & -\\

Noun phrase rate & CoreNLP (6) & $4.89\pm1.64$ & - & - & - & - & - & - & - & - & - & - & - & -\\

Number of verb phrases$\dagger$ & CoreNLP (6) & $302\pm234$ & $\downarrow$\cite{boschi2017connected} & *\cite{boschi2017connected} & - & $\downarrow$\cite{boschi2017connected} & - & - & - & - & - & - & - & - \\

Verb phrase rate & CoreNLP (6) & $3.31\pm1.23$ & - & - & - & - & - & - & - & - & - & - & - & -\\

Number of infinitive phrases$\dagger$ & CoreNLP (6) & $16.2\pm12.7$ & - & - & - & $\downarrow$\cite{boschi2017connected} & - & - & - & - & - & - & - & -\\

Infinitive phrase rate & CoreNLP (6) & $0.189\pm0.120$ & - & - & - & - & - & - & - & - & - & - & - & -\\

Number of prepositional phrases$\dagger$ & CoreNLP (6) & $89.0\pm74.4$ & - & - & - & $\downarrow$\cite{boschi2017connected} & - & -  & - & - & - & - & - & -\\

Prepositional phrase rate & CoreNLP (6) & $0.979\pm0.470$ & - & - & - & - & - & - & - & - & - & - & - & -\\

Number of dependent clauses$\dagger$ & CoreNLP (6) & $72.8\pm67.5$ & *\cite{boschi2017connected} & *\cite{boschi2017connected} & *\cite{boschi2017connected} & $\downarrow$\cite{boschi2017connected} & - & *\cite{boschi2017connected} & $\downarrow$\cite{boschi2017connected} & *\cite{boschi2017connected} & - & - & - & -\\

Dependent clause rate & CoreNLP (6) & $0.780\pm0.400$ & - & - & - & - & - & - & - & - & - & - & - & -\\

Max Yngve depth & CoreNLP (6) & $4.98\pm0.707$ & *\cite{boschi2017connected} & *\cite{boschi2017connected} & - & *\cite{boschi2017connected} & - & - & - & - & - & - & -  & -\\

Mean Yngve depth & CoreNLP (6) & $2.69\pm0.345$ & - & - & - & - & - & - & - & - & - & - & - & -\\

Total Yngve depth & CoreNLP (6) & $58.0\pm27.6$ & - & - & - & - & - & - & - & - & - & - & - & -\\

Parse tree height & CoreNLP (6) &  $10.4\pm1.71$ & *\cite{boschi2017connected} & *\cite{boschi2017connected} & - & *\cite{boschi2017connected}& - & - & - & - & - & - & - & -\\ \hline

 \textbf{Discourse \& pragmatic features} &   & &  &  &  &   &  &  &  & & & & & \\
 \hline
 Number of discourse markers$\dagger$ & Stanza (66) & $75.6\pm62.7$ & $\uparrow$\cite{boschi2017connected} & - & - & - & - & - & - & - & - & - & - & -\\

Discourse marker rate & Stanza (66) & $0.834\pm0.465$ & - & - & - & - & - & - & - & - & - & - & - & -\\

\hline 
\end{tabular}}
\label{tab:table1} 
\end{table*}


\section{BlaBla Architecture} 
\label{section:architecture}

Stanza utilizes deep neural networks for extracting linguistic structures, including the dependency parse tree and POS tags \cite{qi2020stanza} but notably lacking the constituency parse tree, which is used to identify phrase-level structures such as noun phrases and verb phrases. This can be extracted by CoreNLP, which instead employs traditional statistical models \cite{manning2014stanford}. BlaBla processes this low-level information from Stanza and CoreNLP to build a set of linguistic features tailored for the clinical research community. BlaBla takes as input a piece of text or a timestamp-aligned transcript and outputs a CSV file or a Pandas \texttt{DataFrame} of the derived features; timestamp-aligned transcripts are required only to calculate phonetic and phonological features. Language support for each feature is determined by its dependence on Stanza and/or CoreNLP. The languages supported by CoreNLP are English, Arabic, Chinese, French, German and Spanish \cite{manning2014stanford}, whereas Stanza supports 66 languages \cite{qi2020stanza}. BlaBla itself has been tested on the 6 languages supported by CoreNLP and the remaining 60 languages are subject to ongoing validation. For a comprehensive account of language support for each feature, see Table \ref{tab:table1}. 

We release BlaBla under the GNU GPL v3 open-source license to grant its users the freedom to use, adapt and build on top of this library. We hope that this will encourage contributors from both research and industry to offer their time, knowledge and support to the future of this project.

\subsection{Motivation}
The features implemented in BlaBla are chosen to reflect the most up-to-date feature sets commonly used in the diagnosis of neurodegenerative, motor and affective conditions. This was achieved through a clinical literature review drawing on a number of recent review papers \cite{boschi2017connected,low2020automated,tomik2010dysarthria,noffs2018speech,chan2019speech}. The selected features were further split into four categories: \emph{phonetic and phonological} (including features such as the hesitation ratio and the speech rate), \emph{lexicosemantic} (including the noun rate and idea density), \emph{morphosyntactic and syntactic} (including the proportion of auxiliary verbs and the mean Yngve depth) and \emph{discourse and pragmatic} (including the rate of discourse markers), following the nomenclature of \cite{boschi2017connected}. Further details of the features and their clinical motivation can be found in Table \ref{tab:table1}.

Also included in Table \ref{tab:table1} is a set of reference values for each feature derived from manual transcripts from the AMI Meeting Corpus \cite{carletta2005ami}, excluding phonetic and phonological features. A total of 682 single-speaker transcripts were extracted from diarized meeting transcripts; each of these represents one data point in the distributions described by the reference values.

\subsection{Using BlaBla}

 There are two distinct ways to use BlaBla. The first is the native Pythonic way, whereby the user imports and instantiates the \texttt{DocumentProcessor} class with the path to the configuration file and text language. Features can then be extracted from the input via the \texttt{compute\_features} method. This Python interface allows the user to easily integrate BlaBla with other frameworks such as NumPy and TensorFlow. The second way to use BlaBla is via the command line interface (CLI), which extracts features from a text/JSON file or directory of such files. The CLI reads its configuration from a YAML file to determine which features are to be extracted. Sentence parsing can be computationally burdensome and so, to address this, BlaBla's calls to Stanza and CoreNLP are multithreaded and GPU-enabled by default. See BlaBla's documentation for a comprehensive description of usage, structure and functionality.  

\begin{figure*}[!htbp]
     \centering
     \begin{subfigure}{0.45\textwidth}
         \centering
         \includegraphics[width=\textwidth]{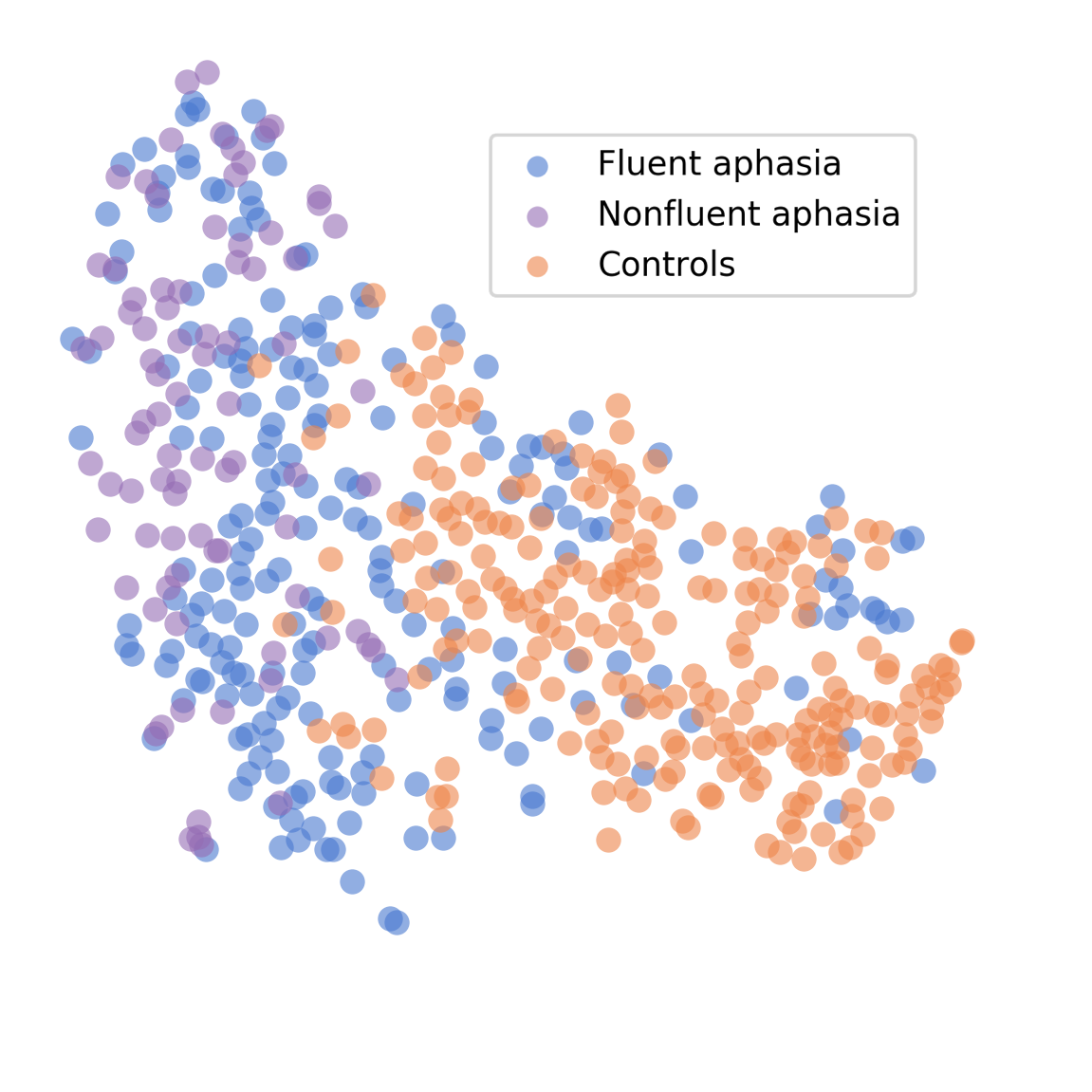}
     \end{subfigure}
     \hfill
     \begin{subfigure}{0.45\textwidth}
         \centering
         \includegraphics[width=\textwidth]{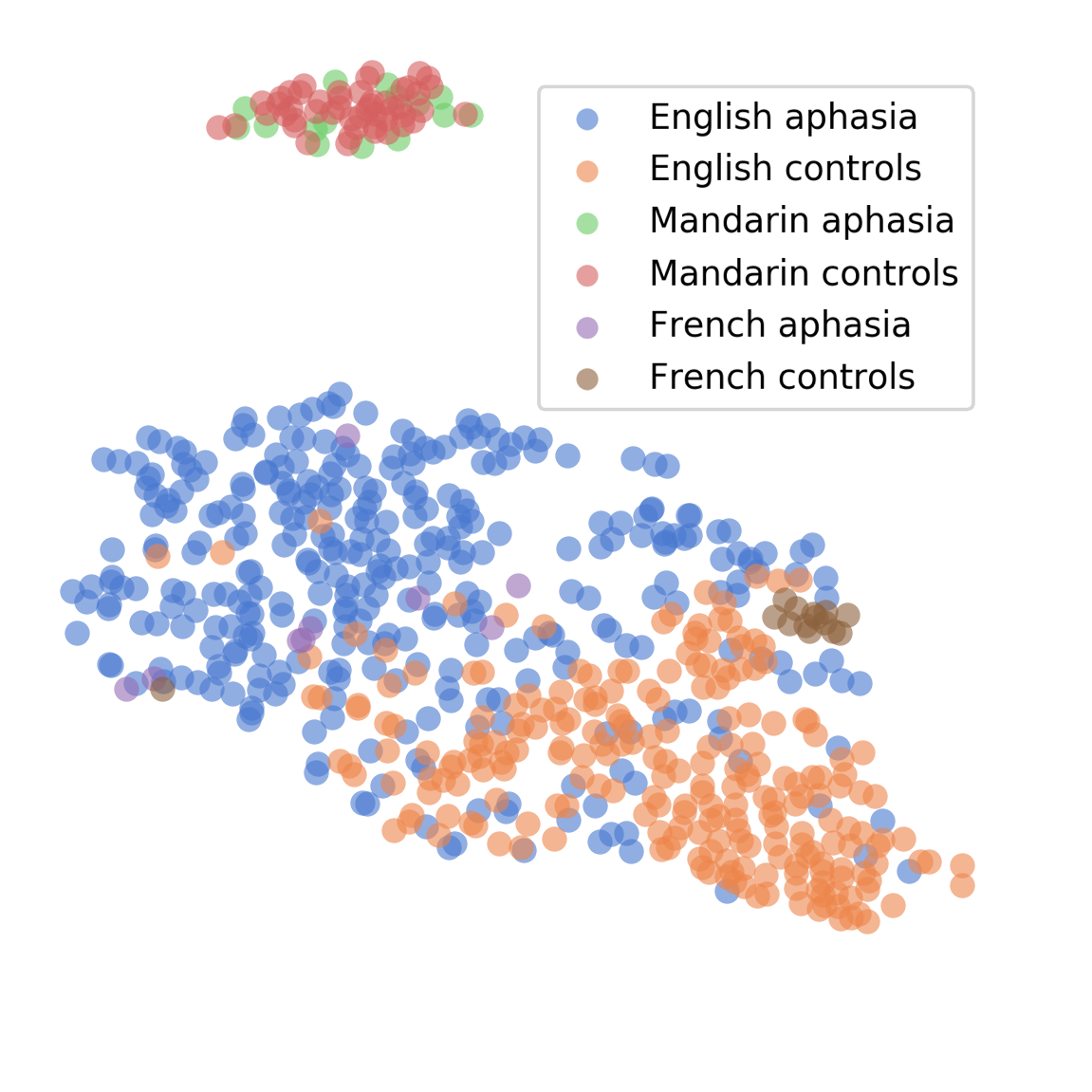}
     \end{subfigure}
     \caption{t-SNE visualizations of a subset of BlaBla features on story narration tasks from English, French and Mandarin corpora. Left: visualization of fluent aphasia, nonfluent aphasia and controls derived from the English corpora only. Right: visualization of aphasic and control patients from corpora spanning English, Mandarin and French.}
     \label{fig:tsne_results}
\end{figure*}

\section{Application: Visualizing and Classifying Aphasia Across Multiple Languages} 

\subsection{Experimental Design}

The features implemented in BlaBla are motivated by the clinical literature, with considered conditions including multiple types of aphasia (see Table \ref{tab:table1}). Aphasia broadly refers to a class of language disorders that leave their sufferer with difficulties understanding or producing language \cite{clark2003aphasia}. The characteristic patterns of impairment can be classified as either fluent or nonfluent aphasias \cite{gordon2020fluent}. Linguistic changes due to aphasia have been the subject of much research and many are well-documented. These include word-finding difficulties and naming disturbances \cite{dressler1988text, larfeuil1997analysis}, changes in noun/verb rates \cite{bates1991noun, hillis2004deterioration}, changes in rates of closed-class words and the omission of determiners \cite{friederici1982syntactic,hofstede1992agrammatic}, and reduced syntactic complexity/accuracy \cite{bird1996cinderella, schwartz1994disordered}. These features can be used for classification of aphasias; in \cite{fraser2014automated}, for instance, the authors extract 58 linguistic measures (such as the inflected verb rate, the height of the parse tree and the rate of demonstratives) from the widely-used \emph{Cinderella} story narration task \cite{bird1996cinderella} to classify primary progressive aphasia (PPA) patients using machine learning. They claim an accuracy of 100\% distinguishing between fluent PPA and healthy controls, and 75\% distinguishing between fluent PPA and nonfluent PPA on their small dataset.

In this section, we take the example of dimensionality reduction of linguistic features for aphasic patients and healthy controls in three languages to illustrate the clinical relevance of BlaBla features. We use the AphasiaBank dataset \cite{macwhinney2011aphasiabank}, a corpus of clinically diagnosed aphasic participants and healthy controls containing manually transcribed and annotated transcripts. We extract the \emph{Cinderella} story narration task from the combined English corpora and the French corpus, along with the \emph{Cry Wolf} story narration task from the Mandarin corpus, where the protocol was adapted for cultural reasons. Extraction of annotation-free transcripts was performed using the CLAN tool \cite{macwhinney2010transcribing} and all linguistic analysis was performed using BlaBla v0.1. For the fluent aphasic, nonfluent aphasic and control participants in the English corpora, we take all 43 non-phonetic and -phonological BlaBla features from Table \ref{tab:table1} and visualize them using a 2D t-SNE to visually identify groups of clinical significance. Only a subset of these features has been previously identified with aphasia but most have not been explicitly studied, so we choose to include them all for illustrative purposes. We repeat this procedure for aphasic participants versus controls in each of the three languages using a restricted feature list that excludes directly length-dependent features (those indicated by a dagger in Table \ref{tab:table1}) and features which are undefined for Mandarin (those relating to verb forms) to facilitate a fairer comparison between different languages and tasks.


English tends to dominate clinical language datasets \cite{forbes2012aphasiabank,gratch2014distress,becker1994natural}, so understanding the extent to which feature-based linguistic analysis generalizes across languages is of great interest. We train a simple aphasia/control classifier using a linear SVM on an English train set and validate it on English, French and Mandarin test sets as a first step towards this. To encourage robustness to task and language variation, we again use the restricted feature list. To account for data scarcity and feature redundancy, we first select the five most important features through recursive feature elimination and use only these to perform the classification. For simplicity, we randomly select balanced train and test sets. 


\subsection{Results}

The results of the t-SNE visualizations are shown in Figure \ref{fig:tsne_results}. Despite being fully unsupervised, the distinction between the controls and nonfluent aphasic participants in English is clear. Fluent aphasia is visually harder to separate from both nonfluent aphasia and controls, which we note does not necessarily imply that these groups are difficult to separate in their full 43-dimensional space. Data from the French corpus appears largely consistent with the English corpora, whereas the Mandarin corpus appears less so. This could be due to innate linguistic differences or those arising from the task variation, though scarcity of data should be noted before drawing conclusions.

\begin{table}[ht]
\caption{Performance of the aphasia/control classifier on the English, French and Mandarin test sets.}
\centering
\begin{tabular}{c c c c}
\hline
& English & French & Mandarin  \\ [0.5ex]
\hline
Train $N_\mathrm{aph}:N_\mathrm{contr}$ & 201:201 & - & -  \\
Test $N_\mathrm{aph}:N_\mathrm{contr}$ & 36:36 & 9:9 & 15:15  \\
Accuracy & 0.903 & 0.833 & 0.500  \\
F1 score & 0.902 & 0.833 & 0.333 \\
\hline 
\end{tabular}
\label{tab:intercentroid}
\end{table}

Table \ref{tab:intercentroid} shows the results of the aphasia/control classifier trained on English and tested on each of the three languages. The five features chosen through recursive feature elimination were: the noun-verb ratio ($\uparrow$), the pronoun-noun ratio ($\uparrow$), the mean Yngve depth ($\downarrow$), the pronoun rate ($\downarrow$) and the content density ($\downarrow$), where the arrows indicate the change direction indicating aphasia as implied by the linear SVM coefficient sign. The classifier achieves an accuracy of 90.3\% on the English test set and 83.3\% on the French test set but does not perform better than random on the Mandarin test set. The results suggest that this simple approach may generalize better to other European languages than more distant languages. Adoption of domain adaptation techniques \cite{pan2010domain} may be necessary to achieve the cross-language and -task generalizability required here. Other complicating factors include the small size of the data sets and the differing composition of aphasia types within the `aphasia' class. Finally, we note that we chose not to optimize model parameters but rather to use this example to illustrate the ease of multilingual analysis using BlaBla.



\section{Conclusions}

In this paper we introduced BlaBla, a novel Python package for NLP with application to the clinical domain. The architecture and usage were described along with a summary of the features available, their clinical relevance and reference values. Aphasic discourse was used to illustrate the applicability and generalizability of the feature set, which we plan to expand significantly in the future. We hope that BlaBla will facilitate the next wave of ML-powered clinical NLP analyses and prove to be an invaluable and evolving resource for both experts and newcomers to the field.

\section{Acknowledgements}

The authors would like to acknowledge and thank the participants, organizers and sponsors who contribute to and maintain the AphasiaBank\footnote{https://aphasia.talkbank.org/} database, without whom the aphasia illustration used in this paper would not have been possible.

\bibliographystyle{IEEEtran}

\bibliography{mybib}

\end{document}